\documentclass[12pt,a4paper]{article}
\usepackage[margin=25mm]{geometry}
\usepackage{graphicx,amsmath,amssymb,amsfonts,bm,booktabs,multirow}
\usepackage[numbers,sort&compress]{natbib}
\usepackage{caption,longtable}
\usepackage[section]{placeins}

\usepackage[hidelinks]{hyperref}
\renewcommand{\baselinestretch}{1.15}
\hypersetup{pdftitle={Structural recovery of multiaxis inertial signals during short-term saturation},pdfauthor={Haoyang Wang; Junjie Yi; Jingyue Zhang; Jiale Gong},pdfsubject={}}
\begin{document}
\begin{center}
\setlength{\parskip}{0pt}
{\Large\bfseries Structural recovery of multiaxis inertial signals during short-term saturation\par}
\vspace{0.7em}
\begingroup
\renewcommand{\baselinestretch}{1.1}
{\fontsize{11}{13}\selectfont Haoyang Wang, Junjie Yi, Jingyue Zhang, Jiale Gong\raisebox{0.45ex}{\fontsize{8}{8}\selectfont *}\par}
\vspace{0.5em}
{\fontsize{10}{12}\selectfont The Future Laboratory, Tsinghua University,\par
Haidian District, Beijing 100084, China\par}
\vspace{0.5em}
{\fontsize{10}{12}\selectfont
Email addresses: \href{mailto:vwang6925@gmail.com}{vwang6925@gmail.com} (H. Wang); \href{mailto:gongjiale@mail.tsinghua.edu.cn}{gongjiale@mail.tsinghua.edu.cn} (J. Gong)\par
\vspace{0.2em}
\raisebox{0.45ex}{\fontsize{8}{8}\selectfont *}Corresponding author: Jiale Gong.\par}
\endgroup
\end{center}
\begin{abstract}
Sensing highly dynamic motion requires continuous inertial measurements, which are limited by the sensor range. Brief saturation clips the peaks and obscures the waveform of an inertial component, while other axes may continue to record the motion. We propose Geo, a geometric recovery model that relates short-term saturation to the local structure of a multiaxis inertial trajectory. Under inertial continuity and multiaxis geometric consistency, Geo expresses target-component increments in terms of displacement along the observable projection, provided that the auxiliary channels remain observable and the local spatial relation persists. SlopeGeo implements this relation as an online estimator, combining pre-event secant slopes with current auxiliary-plane displacement and aggregating candidate estimates through a validation-weighted median. A local error bound quantifies the contributions of slope variation, slope-estimation error and projected chord--arc discrepancy. We test the method by applying controlled software clipping to simulated signals, self-collected human motion and the INSANE and UZH-FPV flight datasets. Relative to clipped input, SlopeGeo reduces pooled overrange root-mean-square error, normalized by the clipping threshold, by 54.6\%, 20.6\%, 23.1\% and 6.7\%, respectively, achieving the lowest pooled error among the seven compared methods for each source. Geo provides a geometric basis for recovering missing measurements from existing inertial channels and extending the usable measurement range during highly dynamic motion.
\end{abstract}

\section{Introduction}

Motion sensing supports agile aircraft, legged robots and wearable human--machine systems \cite{ref01,ref02,ref03,ref36,ref37}. Sensor structure and operating mode jointly determine measurement range, sensitivity, noise and bandwidth \cite{ref04,ref05,ref06,ref07,ref08,ref38}. During highly dynamic motion, an inertial signal may exceed the sensor range, leaving a plateau in the recorded waveform. The peak amplitude and changes within the peak are no longer recorded directly, affecting motion estimation and subsequent integration \cite{ref09,ref13,ref16}.

Existing approaches extend measurement range through device design or improve motion estimation using additional observations. Extended-range MEMS structures and topology optimization modify the mechanical response of the sensing element \cite{ref06,ref38}. Redundant gyroscopes improve angular-rate measurement, while geomagnetic observations and model-based smoothing support motion estimation during saturation \cite{ref09,ref12,ref13,ref14,ref16,ref15}. Related motion-sensing studies combine flow and inertial measurements to estimate limb velocity, acceleration and attitude \cite{ref37}. Biomechanics-informed tracking uses recursive state estimation to incorporate coupling between knee degrees of freedom and muscle--tendon constraints at the hip and ankle \cite{ref39}. Learning methods address gyroscope calibration \cite{ref17}, multimodal motion prediction \cite{ref18} and wearable human--machine interfaces under motion artefacts \cite{ref36}. 

These studies address sensor performance and motion estimation through device design, additional sensing, biomechanical constraints or learned mappings. The measurement problem considered here is the recovery of a clipped component waveform from the remaining observations within the same IMU. The central question is which local relation connects the observed components to the missing increment, and under what conditions that relation supports reconstruction. Geo formulates this relation geometrically: displacement along the observable projection provides a coordinate for the missing component, whose local slope is estimated from reliable pre-saturation measurements. The resulting model specifies the recovery conditions and relates reconstruction error to changes in the local trajectory.

\begin{figure}[!htbp]
\centering
\includegraphics[width=0.96\textwidth,height=0.60\textheight,keepaspectratio]{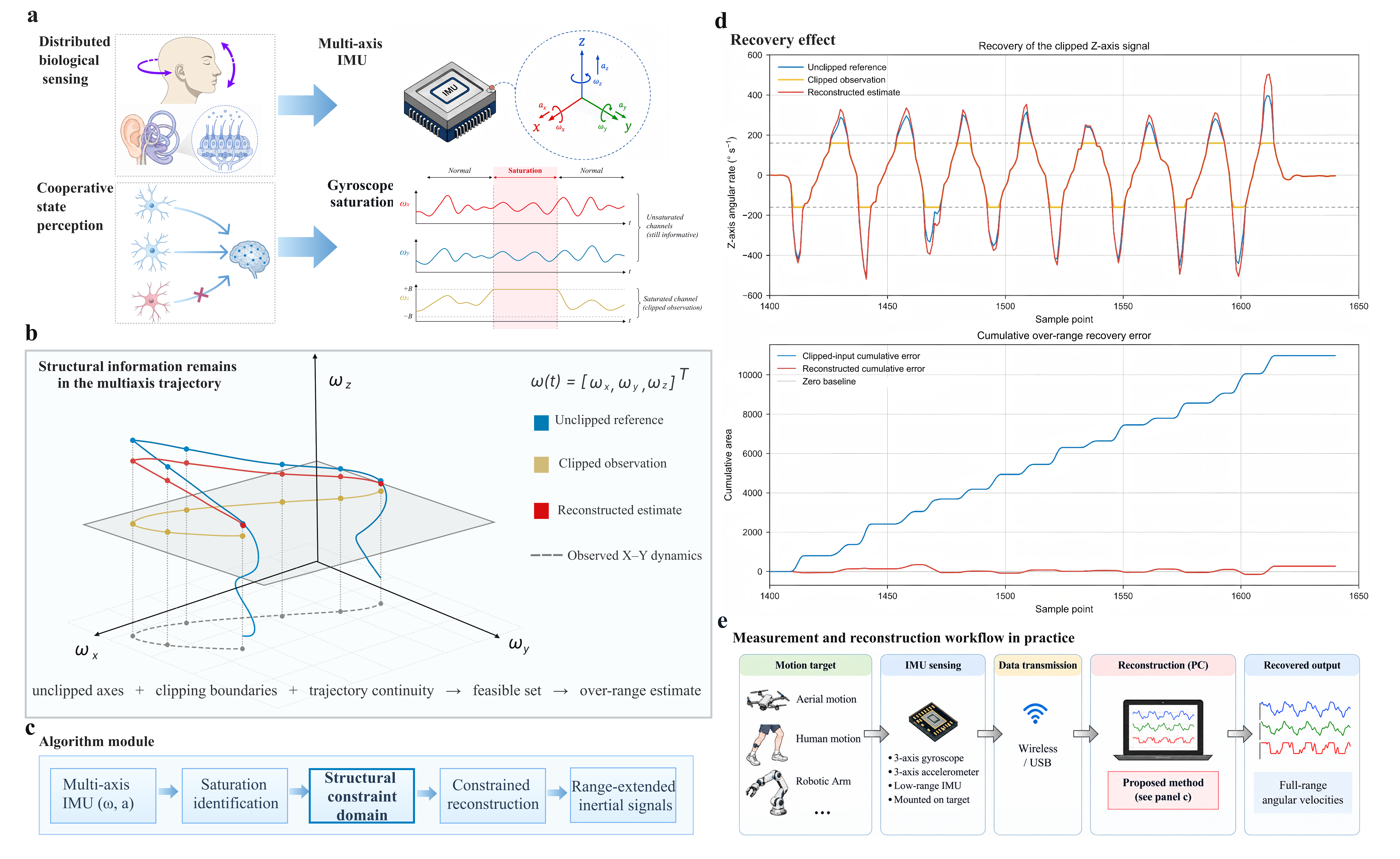}
\caption{\textbf{From distributed biological sensing to structural recovery in a multiaxis inertial system.} \textbf{a,} Conceptual correspondence between distributed biological sensing and a multiaxis IMU: multiple response channels jointly encode motion, while a three-axis gyroscope records the angular velocity of one rigid body. \textbf{b,} Local trajectory in angular-velocity component space. The $z$ axis is clipped at $\pm B$, whereas the $x$--$y$ projection remains observable and constrains the missing change; the legend distinguishes the unclipped reference, clipped observation, reconstructed estimate and observed $x$--$y$ dynamics. \textbf{c,} Processing steps: multiaxis acquisition, saturation detection, definition of the structural-constraint domain, constrained reconstruction and inertial output beyond the original range. \textbf{d,} Example waveform showing the unclipped reference, clipped observation and reconstructed estimate, together with their cumulative overrange error. \textbf{e,} Measurement--reconstruction workflow for aerial, human and robotic motion. The schematic shows the structural constraint and application workflow. Online SlopeGeo reconstruction uses only target-axis history before saturation and measurements from currently available unsaturated channels.}
\label{fig:overview}
\end{figure}

Population coding, in which information is extracted from multiple response channels, provides conceptual inspiration \cite{ref19,ref20,ref21,ref22,ref23,ref24}. Angular-velocity and acceleration measurements each form a trajectory in their respective component spaces. This multiaxis representation provides the geometric setting for Geo (Fig.~\ref{fig:overview}a,b). When the $z$ component is clipped at $\pm B$, the available $x$--$y$ observations retain the planar projection of the trajectory. This projection describes how the trajectory progresses during saturation. Together with the local cross-axis relation, it constrains changes in the missing component. Reconstruction uses the spatial slope estimated immediately before clipping, assuming that it remains applicable over the brief saturation event.

Inertial continuity, multiaxis geometric consistency and physical feasibility define the Geo recovery conditions. SlopeGeo implements this relation using a pre-event anchor and secant slopes estimated over different history lengths. Each slope converts the current displacement in the auxiliary plane into a candidate estimate of the missing component. Reliable samples held out for validation determine the weights in the weighted median (Fig.~\ref{fig:overview}c--e). A local error analysis relates reconstruction accuracy to slope variation, slope-estimation error and projected-path geometry. In this study, the mathematical derivation and quantitative validation focus on angular velocity. We evaluate Geo through SlopeGeo using simulated signals, self-collected human motion and the INSANE and UZH-FPV flight datasets, with controlled software clipping retaining the complete waveforms as references. 

\clearpage
\section{Geometric reconstruction method}\label{sec:methods}

When an inertial component saturates, its missing amplitude must be estimated from measurements that remain available. Geo describes their relation through the local trajectory formed by the signal components. We formulate the physical assumptions and geometric relation, construct SlopeGeo from reliable history and current auxiliary measurements, and derive the local reconstruction error bound.

\subsection{Observable axes preserve local trajectory information}\label{sec:phenomenon}

\begin{figure}[!htbp]
\centering
\includegraphics[width=0.96\textwidth,height=0.40\textheight,keepaspectratio]{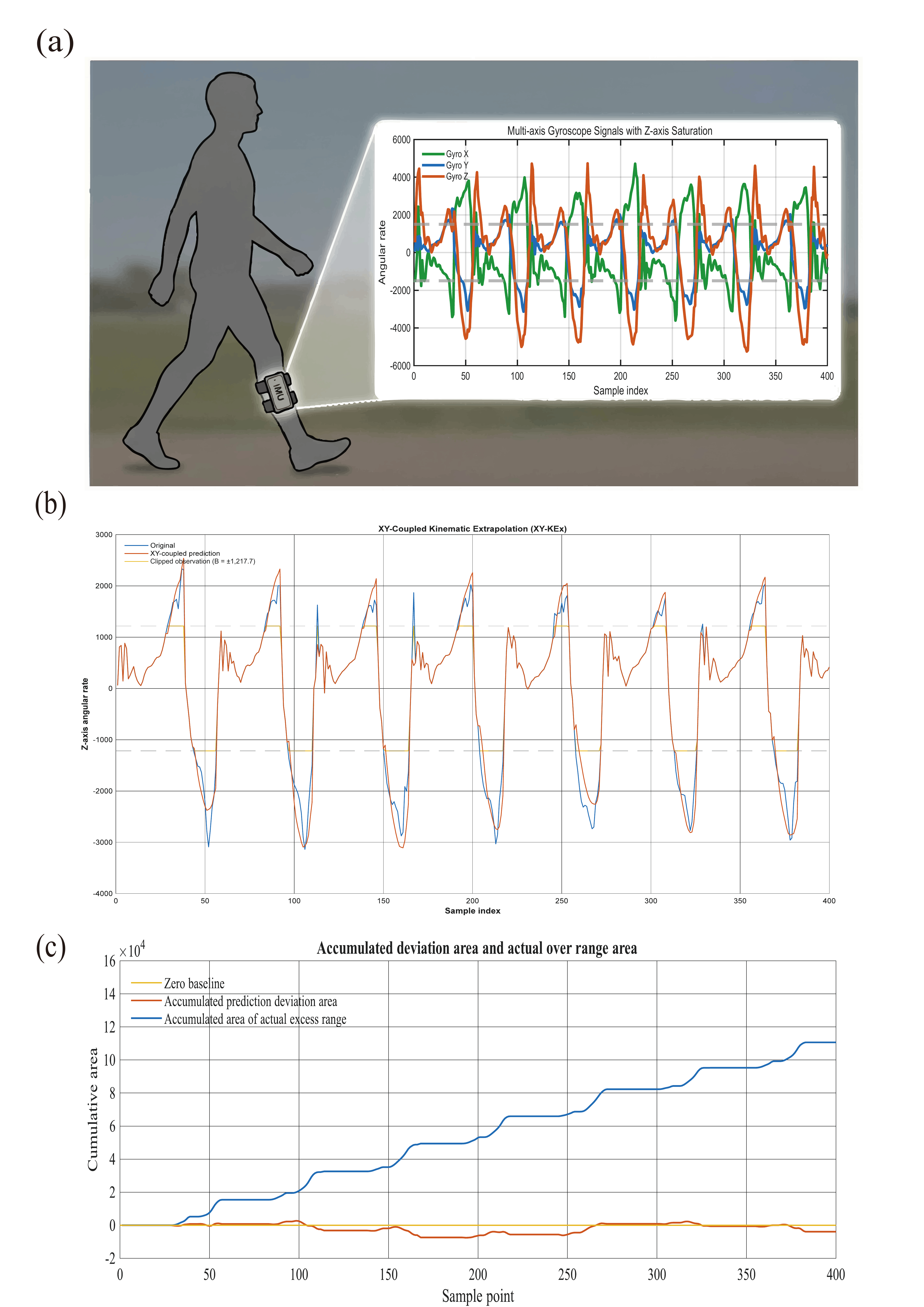}
\caption{\textbf{Local trajectory information during clipping of a target component.} A walking example from HuGaDB~\cite{ref25} shows the target-axis waveform, auxiliary-axis response and local structural reconstruction. This example illustrates the spatial relation evaluated using the four data sources below.}
\label{fig:phenomenon}
\end{figure}

HuGaDB is a wearable inertial dataset of human activities, including locomotion~\cite{ref25}. We use a walking example to examine the local trajectory when a gyroscope component is clipped over a finite interval (Fig.~\ref{fig:phenomenon}). Clipping produces a plateau in the target component, but the auxiliary components continue to vary and retain the planar projection of the local trajectory. The reconstructed signal follows the overall waveform and its trajectory in angular-velocity space. This example motivates using the auxiliary axes to define a spatial coordinate for recovery. Section~\ref{sec:results} evaluates this approach across four data sources using all evaluated overrange samples.

\FloatBarrier
\subsection{Signal model and saturation formulation}\label{sec:signal-model}

At discrete time $k$, the true angular velocity is
\begin{equation}
\bm{\omega}(k)=\begin{bmatrix}\omega_x(k)&\omega_y(k)&\omega_z(k)\end{bmatrix}^{\mathsf T}.
\label{eq:omega-vector}
\end{equation}
For axis $i\in\{x,y,z\}$, let $y_i(k)$ denote the pre-clipping measurement and $\nu_i(k)$ its equivalent measurement perturbation. The observable clipped output is
\begin{equation}
 y_i(k)=\omega_i(k)+\nu_i(k),\qquad
 \widetilde{\omega}_i(k)=\operatorname{sat}_{B_i}\!\left(y_i(k)\right),
\label{eq:measurement-model}
\end{equation}
where $B_i>0$ is the clipping boundary. The saturation operator is
\begin{equation}
\operatorname{sat}_{B_i}(u)=
\begin{cases}
-B_i,&u<-B_i,\\
u,&|u|\le B_i,\\
B_i,&u>B_i.
\end{cases}
\label{eq:saturation-operator}
\end{equation}
We formulate the problem for a target component (denoted below by $z$) that saturates over a finite interval while the auxiliary components remain continuously observable within range. Relabelling the axes gives the formulation for another target component. Recovery requires observable auxiliary channels and a spatial relation that remains valid locally.

\subsection{Physical constraints underlying structural recoverability}\label{sec:physical-constraints}

Starting from rigid-body rotational dynamics, we identify the conditions under which the local trajectory can be continued through a short clipping interval. Inertial continuity describes its local regularity (H1); an observable projection and a persistent cross-axis slope define the geometric relation (H2); application-dependent magnitude bounds describe physical admissibility (H3). The derivation below distinguishes the consequences of the dynamics from the additional local assumptions used for reconstruction.

\subsubsection{Inertial continuity (H1)}\label{sec:h1}

With a fixed positive-definite inertia matrix, bounded angular velocity and bounded external torque, the angular-velocity trajectory has a bounded local rate of change and is therefore locally Lipschitz continuous:
\begin{equation}
\|\bm{\omega}(t+\tau)-\bm{\omega}(t)\|\le L_{\omega}|\tau|.
\label{eq:h1-lipschitz}
\end{equation}
To obtain the second-order local remainder below, we also assume that $\mathbf F(\bm{\omega}(t),\bm{\tau}(t))$ is locally Lipschitz along the trajectory. This regularity assumption supports short-time continuation of the trajectory while allowing angular speed to vary.

\subsubsection{Multiaxis dynamic and geometric consistency (H2)}\label{sec:h2}

Rigid-body rotation satisfies~\cite{ref28}
\begin{equation}
\mathbf J\dot{\bm{\omega}}+\bm{\omega}\times(\mathbf J\bm{\omega})=\bm{\tau},
\label{eq:euler}
\end{equation}
or
\begin{equation}
\dot{\bm{\omega}}=\mathbf F(\bm{\omega},\bm{\tau})=
\begin{bmatrix}F_x&F_y&F_z\end{bmatrix}^{\mathsf T}.
\label{eq:state-form}
\end{equation}
In the principal-axis frame,
\begin{align}
\dot{\omega}_x&=\frac{J_y-J_z}{J_x}\omega_y\omega_z+\frac{\tau_x}{J_x},\nonumber\\
\dot{\omega}_y&=\frac{J_z-J_x}{J_y}\omega_z\omega_x+\frac{\tau_y}{J_y},\nonumber\\
\dot{\omega}_z&=\frac{J_x-J_y}{J_z}\omega_x\omega_y+\frac{\tau_z}{J_z}.
\label{eq:euler-components}
\end{align}
At fixed torque, for example,
\begin{equation}
\frac{\partial\dot{\omega}_z}{\partial\omega_x}=\frac{J_x-J_y}{J_z}\omega_y.
\label{eq:cross-axis-derivative}
\end{equation}
For unequal principal moments and non-zero auxiliary angular velocity, this derivative describes cross-axis coupling at fixed torque. If the sensor axes are not aligned with the principal inertia axes, the coupling remains described by the general inertia matrix in Eq.~\ref{eq:euler}.

Define the projected arc length in the observable $xy$ plane as
\begin{equation}
s_{xy}(t;t_0)=\int_{t_0}^{t}\sqrt{\dot{\omega}_x^2(\xi)+\dot{\omega}_y^2(\xi)}\,\mathrm d\xi.
\label{eq:projected-arclength}
\end{equation}
When the rate of motion along the projected trajectory satisfies $v_{xy}(t)=\sqrt{\dot{\omega}_x^2+\dot{\omega}_y^2}\ge v_{\min}>0$ on the local interval, this projection is non-degenerate and
\begin{equation}
\frac{\mathrm d\omega_z}{\mathrm ds_{xy}}=
\frac{\dot{\omega}_z}{\sqrt{\dot{\omega}_x^2+\dot{\omega}_y^2}}\equiv g(t).
\label{eq:spatial-slope}
\end{equation}
Integrating along the projected path gives
\begin{equation}
\omega_z(t)-\omega_z(t_0)=\int_{s_{xy}(t_0)}^{s_{xy}(t)}g(s)\,\mathrm ds.
\label{eq:continuous-path-relation}
\end{equation}
The positive projected speed makes $s_{xy}$ strictly increasing, so $g(s)$ denotes $g(t(s))$ along this time-ordered path. The relation parameterizes the missing component by accumulated path length; it does not require a unique mapping from planar position to the target value. Recovery additionally assumes that the pre-saturation slope persists locally, with its variation quantified below. This assumption supplies the continuation model beyond the rotational dynamics described by the Euler equation. Curvature or a return towards the anchor increases the difference between path length and endpoint distance, which enters the reconstruction error bound.

\subsubsection{Discrete spatial relation}\label{sec:discretization}

A first-order discretization over sampling interval $\Delta t$ gives
\begin{equation}
\Delta\bm{\omega}(k)=\mathbf F(k-1)\Delta t+O(\Delta t^2),
\label{eq:first-order-discretization}
\end{equation}
where $\Delta\omega_i(k)=\omega_i(k)-\omega_i(k-1)$. Define
\begin{equation}
\Delta s_{xy}(k)=\sqrt{[\Delta\omega_x(k)]^2+[\Delta\omega_y(k)]^2}.
\label{eq:single-step-projection}
\end{equation}
For a locally non-degenerate projection, the common time increment can be eliminated to yield
\begin{equation}
\Delta\omega_z(k)=g(k-1)\Delta s_{xy}(k)+O(\Delta t^2),
\label{eq:discrete-spatial-relation}
\end{equation}
with
\begin{equation}
g(k-1)=\frac{F_z(k-1)}{\sqrt{F_x^2(k-1)+F_y^2(k-1)}}.
\label{eq:discrete-slope}
\end{equation}

\subsubsection{Physical feasibility (H3)}\label{sec:h3}

We assume a local bound on angular-velocity magnitude:
\begin{equation}
\|\bm{\omega}(k)\|\le\Omega_{\max},
\label{eq:physical-bound}
\end{equation}
where $\Omega_{\max}$ is an application-dependent physical bound. If it is known, the observable components imply $|\omega_z|\le\sqrt{\Omega_{\max}^2-\omega_x^2-\omega_y^2}$ whenever the radicand is non-negative. This vector constraint characterizes physical admissibility. The evaluated implementation instead uses the observed saturation sign and a scalar numerical cap $3B$ shared by all methods; it neither estimates $\Omega_{\max}$ nor enforces this vector constraint. The cap controls finite output magnitude and is distinct from the physical bound. H1 and H2 provide the regularity and local geometric assumptions used in the error analysis.
\begin{figure}[!htbp]
\centering
\includegraphics[width=0.96\textwidth,height=0.60\textheight,keepaspectratio]{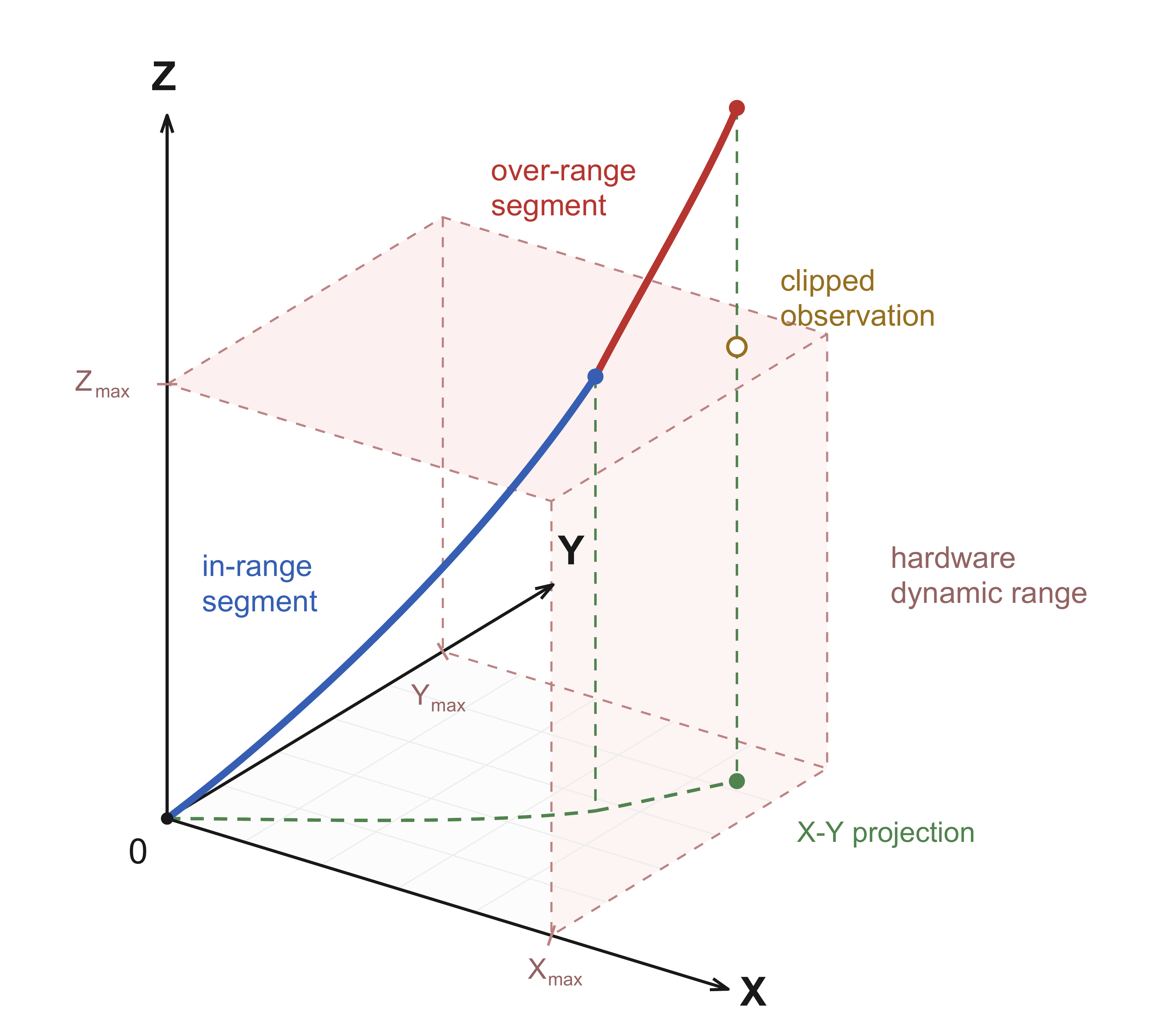}
\caption{\textbf{Physical constraints supporting local recovery of a saturated inertial channel.} The recovery conditions comprise inertial continuity (H1), a non-degenerate observable-plane projection with short-term slope persistence (H2), and application-dependent physical bounds (H3). The implementation uses saturation-sign consistency and a separate scalar numerical cap. The error analysis includes the difference between the projected arc and chord lengths.}
\label{fig:constraints}
\end{figure}

\FloatBarrier
\subsection{Online implementation of Geo through SlopeGeo}\label{sec:reconstruction-method}

SlopeGeo implements the Geo relation online. Before saturation, it obtains the anchor and candidate spatial slopes, using reliable samples held out for validation to set their weights. During saturation, new auxiliary measurements update the projected displacement used to reconstruct the clipped component.

\subsubsection{Saturated intervals and recovery anchors}\label{sec:anchor}

For an event group with start and end indices $k_s$ and $k_e$, the saturated points identified from the clipped output are
\begin{equation}
\mathcal S=\{k\in[k_s,k_e]:|\widetilde{\omega}_z(k)|\ge B-\varepsilon_{\mathrm{sat}}\},\qquad
k_0=k_s-1,
\label{eq:saturated-set}
\end{equation}
where $\varepsilon_{\mathrm{sat}}=\max\{10^{-9},8\,\mathrm{spacing}(B)\}$ is a floating-point boundary tolerance. A run is closed after $G=\max\{2,\operatorname{round}(0.030f_s)\}$ consecutive reliable samples; short gaps are merged into the same group. The observed value $\widetilde{\omega}_z(k_0)$ is the recovery anchor. The history contains only consecutive reliable samples before $k_s$; reconstructed values are not added to it.

\subsubsection{Pre-event slope estimation and consistency selection}\label{sec:slope-estimation}

For each history duration $T_j\in\{0.010,0.015,\ldots,0.295\}\,\mathrm{s}$, the corresponding span in samples is
\begin{equation}
N_j=1+\operatorname{round}(f_sT_j),
\label{eq:history-index}
\end{equation}
with duplicate spans and spans shorter than three samples removed. Each candidate requires $N_j+2$ consecutive reliable samples, including the two validation samples; spans exceeding the available history are omitted. The two endpoints used for fitting are $a_j=k_0-2$ and $b_j=a_j-(N_j-1)=k_0-N_j-1$. The distance between these endpoints in the auxiliary plane gives the pre-saturation projected baseline
\begin{equation}
\rho_{\mathrm{pre},j}=\sqrt{
[\widetilde{\omega}_x(a_j)-\widetilde{\omega}_x(b_j)]^2+
[\widetilde{\omega}_y(a_j)-\widetilde{\omega}_y(b_j)]^2}.
\label{eq:pre-baseline}
\end{equation}
We estimate the pre-saturation slope from the local secant,
\begin{equation}
\widehat g_j=\frac{\widetilde{\omega}_z(a_j)-\widetilde{\omega}_z(b_j)}{\rho_{\mathrm{pre},j}},
\qquad \rho_{\mathrm{pre},j}>\varepsilon_\rho,
\label{eq:slope-estimate}
\end{equation}
where $\varepsilon_\rho=10^{-12}$ in the recorded angular-rate units prevents division by a projected distance close to zero. We reserve the last two reliable samples, $k_0-1$ and $k_0$, for validation. Their Euclidean distance in the auxiliary plane defines $d_{\mathrm{val}}$, and $\Delta z_{\mathrm{val}}=\widetilde\omega_z(k_0)-\widetilde\omega_z(k_0-1)$; candidates with $d_{\mathrm{val}}\le\varepsilon_\rho$ are discarded and the loss for each remaining candidate is
\begin{equation}
L_j=\frac{|\widehat g_jd_{\mathrm{val}}-\Delta z_{\mathrm{val}}|}{B}.
\label{eq:validation-loss}
\end{equation}

\subsubsection{Within-event closed-form recovery and weighted fusion}\label{sec:closed-form}

For a point $k\in\mathcal S$, the projected chord length from the anchor to the current point is
\begin{equation}
\rho_k=\sqrt{
[\widetilde{\omega}_x(k)-\widetilde{\omega}_x(k_0)]^2+
[\widetilde{\omega}_y(k)-\widetilde{\omega}_y(k_0)]^2}.
\label{eq:current-chord}
\end{equation}
Let $L_k$ be the corresponding accumulated projected arc length, so that
\begin{equation}
L_k=\rho_k+\delta_{c,k},\qquad \delta_{c,k}\ge0.
\label{eq:chord-arc-gap}
\end{equation}
Under H2, $g$ is approximated by each pre-event candidate $\widehat g_j$ over a sufficiently short interval. For a locally straight projected trajectory, $L_k\approx\rho_k$, producing the candidate estimate
\begin{equation}
u_j(k)=\widetilde{\omega}_z(k_0)+\widehat g_j\rho_k.
\label{eq:closed-form-recovery}
\end{equation}
We then constrain the candidate to have the sign of the saturated observation,
\begin{equation}
v_j(k)=\sigma_k\max\{|u_j(k)|,B\},\qquad \sigma_k=\operatorname{sign}[\widetilde{\omega}_z(k)].
\label{eq:candidate-feasibility}
\end{equation}
This rule gives the estimate the observed sign and a magnitude of at least $B$.

Candidate weights are
\begin{equation}
w_j=\exp[-(L_j-L_{\min})/\lambda_L],\qquad
\lambda_L=\max\{\operatorname{median}(|L-\operatorname{median}(L)|),0.05\},
\label{eq:candidate-weights}
\end{equation}
The weighted median gives the output before the final safeguard,
\begin{equation}
\widehat z_{\mathrm{pre}}(k)=\operatorname{weighted\text{-}median}(\{v_j(k)\},\{w_j\}).
\label{eq:weighted-median}
\end{equation}
The weighted median combines estimates from the admissible history spans according to their validation weights. If candidates within an error interval carry more than half of the total weight, the median lies within that interval. A single candidate with more than half of the total weight can therefore determine the output.
If no finite candidate has positive weight, we return the saturation boundary with the observed sign. Outside $\mathcal S$, the original in-range measurements are retained. Event closure determines when a new anchor is established; predictions at time $k$ do not depend on the eventual event end $k_e$.

\subsubsection{Final-output safeguard}\label{sec:output-safeguard}

All methods share the final numerical safeguard
\begin{equation}
\widehat z(k)=
\begin{cases}
\sigma_k(1+Q)B,&\widehat z_{\mathrm{pre}}\text{ finite and }|\widehat z_{\mathrm{pre}}|>(1+Q)B,\\
\sigma_kB,&\widehat z_{\mathrm{pre}}\text{ is non-finite},\\
\widehat z_{\mathrm{pre}}(k),&\text{otherwise},
\end{cases}
\qquad Q=2.
\label{eq:q-protection}
\end{equation}
The final magnitude limit is therefore $3B$, applied after candidate selection and prediction. The comparison implementations also align finite saturated-point outputs with the observed sign and enforce a minimum magnitude $B$, as specified in Section~\ref{sec:metrics}.
\subsection{Short-time approximation and recoverability}\label{sec:boundary-method}

Reconstruction uses the pre-event slope throughout the clipping interval. We quantify how changes in the slope and curvature of the projected trajectory affect a candidate estimate, then relate the interval duration to sampling and clipping severity.

At a current event sample $k$, let $m=k-k_0$ be the number of sampling intervals since the anchor. The elapsed duration is
\begin{equation}
T_{\mathrm{sat}}=m\Delta t,
\label{eq:saturation-duration}
\end{equation}
and the accumulated first-order discretization remainder scales as
\begin{equation}
O(m\Delta t^2)=O(T_{\mathrm{sat}}\Delta t).
\label{eq:accumulated-discretization}
\end{equation}
Writing $g_0=g(t_0)$, if the local slope satisfies
\begin{equation}
|g(t)-g_0|\le L_g|t-t_0|
\label{eq:slope-lipschitz}
\end{equation}
and projected speed is bounded by $v_{xy,\max}$, denote the noise-free arc and chord lengths by $L_k^*$ and $\rho_k^*$ and define
\begin{equation}
\varepsilon_g=|g_0-\widehat g_0|,\qquad \delta_{c,k}^*=L_k^*-\rho_k^*.
\label{eq:boundary-errors}
\end{equation}
For one candidate, define the ideal estimate $u^*(k)=\omega_z(k_0)+\widehat g_0\rho_k^*$, where $\rho_k^*$ is the chord of the noise-free auxiliary trajectory, and let $e_k=u^*(k)-\omega_z(k)$. The continuous path identity gives the exact decomposition
\begin{equation}
e_k=(\widehat g_0-g_0)L_k^*-\widehat g_0\delta_{c,k}^*-\int_{t_0}^{t_k}[g(t)-g_0]v_{xy}(t)\,\mathrm dt.
\label{eq:error-decomposition}
\end{equation}
Bounding the integral by $L_gv_{xy,\max}\int_0^{T_{\mathrm{sat}}}t\,\mathrm dt$ yields
\begin{equation}
|e_k|\le\frac{1}{2}L_gv_{xy,\max}T_{\mathrm{sat}}^2
+\varepsilon_gL_k^*+|\widehat g_0|\delta_{c,k}^*
+C_dT_{\mathrm{sat}}\Delta t,
\label{eq:error-bound}
\end{equation}
Here $C_d$ bounds the accumulated discretization term when the discrete spatial relation is used. The terms account for slope variation, slope-estimation error, the difference between arc and chord lengths, and accumulated discretization error. For the continuous path identity the discretization term can be set to zero. For measured data, perturbations in the anchor and chord length add an error bounded by $|\nu_z(k_0)|+|\widehat g_0|\,|\rho_k-\rho_k^*|$; the fitted slope error is already represented by $\varepsilon_g$. This analysis applies to the individual geometric candidate before sign alignment, median aggregation and the final amplitude safeguard. These operations are specified in Eqs.~\ref{eq:candidate-feasibility}--\ref{eq:q-protection}.

For ideal clipping, the target value has the observed sign and satisfies $|\omega_z(k)|\ge B$. Sign alignment $u\mapsto\sigma_k\max\{|u|,B\}$ cannot increase its absolute error. A weighted median with positive weights lies between the smallest and largest aligned candidates. Thus, if $E_j(k)$ bounds the measured error of candidate $j$, the final output obeys
\begin{equation}
|\widehat z(k)-\omega_z(k)|\le\max_{j\in\mathcal J_k}E_j(k)+\bigl(|\omega_z(k)|-3B\bigr)_+,
\label{eq:aggregate-bound}
\end{equation}
where $\mathcal J_k$ contains the retained finite candidates and $(a)_+=\max\{a,0\}$. The additional term vanishes within the numerical cap. If no candidate is available, the boundary output has error $|\omega_z(k)|-B$. This extension connects the local geometric bound to the implemented aggregation and output rules under the stated clipping model.

To relate frequency and clipping severity to duration, consider a locally dominant sinusoid
\begin{equation}
\omega_z(t)=A\sin(2\pi ft+\phi),
\label{eq:local-sinusoid}
\end{equation}
with clipping threshold $B$ and normalized threshold $\beta=B/A$, $0<\beta<1$. Around one positive peak, the duration above threshold is
\begin{equation}
T_{\mathrm{sat}}(f,\beta)=\frac{\pi-2\arcsin\beta}{2\pi f}.
\label{eq:duration-beta}
\end{equation}
Defining over-range severity as $q=(A-B)/A=1-\beta$ gives
\begin{equation}
T_{\mathrm{sat}}(f,q)=\frac{\pi-2\arcsin(1-q)}{2\pi f}.
\label{eq:duration-severity}
\end{equation}
For this sinusoidal model, lower frequency and greater clipping severity lengthen the overrange interval. The range of conditions suitable for reconstruction therefore depends on event duration, the pre-event baseline, local path shape and sampling.
\section{Experimental design}\label{sec:experimental-setup}

We use SlopeGeo to test Geo under different motions and sampling rates. Simulation tests recovery with a known spatial relation; self-collected motion and public flight datasets test recovery from measured signals. All comparisons use the same clipping protocol to assess reconstruction error and cumulative signed error.

\subsection{Evaluation overview}\label{sec:validation}\label{sec:statistics}

We evaluate one 60-s simulated sequence, 32 self-collected motion recordings, 26 INSANE UAV sequences~\cite{ref26} and 31 UZH-FPV sequences~\cite{ref27}. The simulated signal components and their projection are known, allowing a direct test of the spatial relation. The measured recordings cover changes in movement direction and rhythm, different flight environments and high-speed manoeuvres. Each original signal is paired with an input clipped at threshold $B$. Recovery uses pre-event target-axis history and currently available auxiliary measurements; SlopeGeo uses one sampling-rate-dependent configuration across both UAV datasets. The comparison includes clipped input and five recovery baselines, with fixed rules given in Section~\ref{sec:metrics}. All methods share the final-output safeguard $Q=2$, corresponding to $\lvert\hat z\rvert\leq3B$.

Pooled overrange RMSE/$B$ combines normalized squared errors from all evaluated samples. We also calculate the RMSE of each recording, then report the mean and sample standard deviation across recordings, with equal weight for each recording. Sample variance uses the number of recordings minus one as the denominator. For each comparator, we count the recordings in which SlopeGeo has lower RMSE. The simulation contains one realization, so we report its pooled error without estimating variation across recordings. Metric definitions appear in Section~\ref{sec:metrics}. After prediction, we select waveform windows from the reference signals and completed outputs. Each contains two or three clipping runs with good recovery that can be interpreted visually. These windows illustrate waveform shape and cumulative signed error; the dataset summaries include all evaluated overrange samples.

Table~\ref{tab:orders} lists sampling, sample counts and model orders. ARX orders apply to both RLS-ARX and CR-RLS-ARX; orders and delays follow the per-sequence rules in Table~\ref{tab:baselines}.

\begin{table}[!htbp]
\centering\small
\caption{Sampling rates, evaluated samples and baseline model orders.}\label{tab:orders}
\setlength{\tabcolsep}{3pt}
\begin{tabular}{lrrrrrr}\toprule
Source & Records & Overrange & $f_s$ (Hz) & AR & ARX & DMD\\\midrule
INSANE & 26 & 120,645 & 199.881--200.081 & 20 & 8 & 8\\
UZH-FPV & 31 & 240,789 & 199.919--500.036 & 20--50 & 8--20 & 8--20\\
Self-collected & 32 & 3,668 & 5.420--38.462 & 3--4 & 1--2 & 4\\
Simulation & 1 & 12,999 & 600 & 60 & 24 & 24\\
\bottomrule\end{tabular}

\end{table}
\FloatBarrier

\subsection{Controlled simulation}\label{sec:simulation-method}
To test recovery with known shared components and a changing spatial projection, the controlled experiment used one 60-s sequence sampled at $f_s=600$ Hz, with $t_k=k/f_s$ and $k=0,\ldots,N-1$. Two signal components and a time-varying projection were defined as
\begin{equation}
\begin{aligned}
u(t)&=200\sin(2\pi\cdot0.3t)+80\sin(2\pi\cdot2.5t),\\
r(t)&=50\sin(2\pi\cdot0.7t),\\
\phi(t)&=\pi/4+0.2\sin(2\pi\cdot0.5t).
\end{aligned}
\label{eq:simulation-shared}
\end{equation}
\begin{equation}
G_x(t)=r(t),\qquad G_y(t)=u(t)\cos\phi(t),\qquad G_z(t)=u(t)\sin\phi(t),
\label{eq:simulation-clean}
\end{equation}
followed by independent unit Gaussian noise and axis-dependent linear drift,
\begin{equation}
X=G_x+2\epsilon_x+0.02t,\quad Y=G_y+2\epsilon_y+0.016t,\quad Z=G_z+2\epsilon_z+0.004t,\quad \epsilon_i\sim\mathcal N(0,1).
\label{eq:simulation-noise}
\end{equation}
The signal amplitudes and noise standard deviation are expressed in degrees per second, drift slopes in degrees per second squared, and $\phi$ in radians. The target axis is clipped at $B=110^\circ\mathrm{s}^{-1}$ (approximately $1.92$ rad s$^{-1}$). The $u$ component contains 0.3 and 2.5 Hz terms, $r$ has frequency 0.7 Hz, and the projection angle varies at 0.5 Hz. Drift slopes for $X,Y,Z$ are 0.020, 0.016 and 0.004 degrees per second squared, respectively. The online history grid is defined in Eq.~\ref{eq:history-index}. In the reference-assisted sweep, the reference signal is used to select a history span at each frequency and threshold combination. Full-sequence errors are then averaged over four combinations of noise and drift. The reported realization contains 120 saturation events and 12,999 evaluated overrange samples, using the unclipped generated signal as the numerical reference.

\subsection{Recorded motion and clipping protocol}\label{sec:hardware-method}\label{sec:public-data-method}\label{sec:clipping-protocol}

We collected 32 BNO055 IMU recordings of hand-held single-axis reciprocation, in-plane motion, mixed-axis motion and leg movement at different intensities (Fig.~\ref{fig:selfcollected_recovery}a). Thirty-one recordings have a median sampling interval of approximately 26 ms (38.46 Hz); one is sampled at approximately 5.42 Hz. Together they contain 3,668 evaluated overrange samples. Data collection received ethics approval, and all participants provided informed consent. During data curation, the axis with the clearest peaks and largest amplitude excursion was selected as the target and mapped to $z$; the other axes were mapped to $x$ and $y$. This assignment remained fixed during prediction. Clipping and recovery use the recorded gyroscope values directly, with the shared threshold rule defined below. The example in Fig.~\ref{fig:selfcollected_recovery}c uses the recorded $G_z$ channel.

For both public flight datasets, we apply controlled clipping to the original $z$-axis recordings and retain them as references. The evaluation includes 26 INSANE sequences~\cite{ref26} covering indoor, outdoor, indoor--outdoor transition and Mars-analogue UAV flights. Sampling rates range from 199.881 to 200.081 Hz, and the evaluation contains 120,645 overrange samples. Figure~\ref{fig:insane_recovery}a shows the flight platform and trajectory for the displayed example. The evaluation also includes 31 UZH-FPV sequences~\cite{ref27} from high-speed indoor and outdoor drone-racing flights. Sampling rates range from 199.919 to 500.036 Hz, and the evaluation contains 240,789 overrange samples. Figure~\ref{fig:fpv_recovery}a shows the corresponding flight platform and trajectory.

Self-collected and public recordings use the same threshold rule. We detect peaks in the absolute target-axis signal and sort their amplitudes in descending order as $p_{(r)}$. For $M$ peaks, the largest $K=\max\{1,\lceil0.20M\rceil\}$ define $S=(1/K)\sum_{r=1}^{K}p_{(r)}$ and $B=0.70S$. Peaks are local maxima of the unfiltered absolute target-axis signal; for a flat peak, the middle sample is used (rounded down when the plateau has an even number of samples). No additional height, prominence, width or separation criterion is imposed. Only finite peak amplitudes are included. If no local peaks are detected, we apply the same rule to the finite absolute sample values.

Thresholds remain fixed during prediction. We use the reference signal to set the threshold when constructing the dataset and to evaluate the outputs after prediction. Reconstruction uses the clipped target signal and available auxiliary measurements at their original sampling rate, without input filtering. For the public flight records, $f_s$ is the reciprocal of the median timestamp interval and remains fixed for the sequence when converting durations into sample counts. No resampling is performed. The lower-rate self-collected record consequently uses fewer distinct history spans under the same conversion rule.

\subsection{Baseline methods and evaluation metrics}\label{sec:metrics}
The comparison comprises seven methods: SlopeGeo (labelled Ours in the figures), clipped input, Forward AR, RLS-ARX, Hankel-DMD, CR-RLS-ARX and SSMKF-log-S. The methods cover scalar prediction, cross-axis system identification, dynamics in delay coordinates, censored regression and robust state estimation.

Forward AR and RLS-ARX follow autoregressive/system-identification formulations~\cite{ref29}. Hankel-DMD uses delay-embedded dynamic mode decomposition with truncated SVD~\cite{ref30,ref31,ref32}. CR-RLS-ARX applies censored recursive regression~\cite{ref33} to target-axis history and current auxiliary observations. SSMKF-log-S uses a statistical-similarity Kalman framework~\cite{ref34}, updating from reliable measurements and predicting the state during saturation. Tables~\ref{tab:orders} and \ref{tab:baselines} specify the sampling-rate-dependent orders, signal histories, initialization and update rules. All methods share the $Q=2$ final-output safeguard.

The SSMKF-log-S state contains the target signal normalized by $B$ and its temporal derivative. Its transition and observation matrices are $F=\left[\begin{smallmatrix}1&\Delta t\\0&1\end{smallmatrix}\right]$ and $H=[1\;0]$, with process covariance $Q_{\mathrm{KF}}=dd^{\mathsf T}$, $d=[\Delta t^2/2\;\Delta t]^{\mathsf T}$, and measurement variance $R=0.0025$. At a saturated sample, the sign- and magnitude-constrained prediction is written back to the level state before the common final-output cap is applied.

In Table~\ref{tab:baselines}, $f_s$ is in hertz, $\epsilon$ is float64 machine precision, and $\operatorname{round}$ uses Python\textquotesingle s nearest-integer rule. All methods receive unfiltered, unresampled inputs. Saturation is detected with the tolerance in Eq.~\ref{eq:saturated-set}. Sign alignment, the minimum magnitude $B$ and the final cap $3B$ are applied to the outputs; these operations leave internal states unchanged except for the SSMKF-log-S level-state update described above.

\begingroup
\small\setlength{\tabcolsep}{4pt}\renewcommand{\arraystretch}{1.08}
\begin{longtable}{p{0.16\textwidth}p{0.45\textwidth}p{0.33\textwidth}}
\caption{Fixed baseline parameters and update rules.}\label{tab:baselines}\\
\toprule Method & Inputs and parameters & Update rule\\\midrule\endfirsthead
\multicolumn{3}{l}{Table \thetable (continued)}\\
\toprule Method & Inputs and parameters & Update rule\\\midrule\endhead
\midrule\endfoot
\bottomrule\endlastfoot
\raggedright Forward AR & \raggedright Target-axis history. $p=\max\{3,\operatorname{round}(0.100f_s)\}$; $W=\max\{p+3,\operatorname{round}(1.0f_s)\}$. Ridge coefficient $0.001[\operatorname{tr}(D^{\mathsf T}D)/p+\epsilon]$, where $D$ is the regression design matrix. No intercept or mean removal. & \raggedright Fit once at each run onset. Recursive predictions extend the history during saturation.\tabularnewline[6pt]
\raggedright RLS-ARX & \raggedright Target history and current two-axis auxiliary input, all divided by $B$. $p=\max\{1,\operatorname{round}(0.040f_s)\}$; $d=p+3$; forgetting factor $0.995^{200/f_s}$. Initial coefficients are zero and $P=10I$. & \raggedright Update coefficients and covariance at reliable samples. During saturation, predictions advance the state.\tabularnewline[6pt]
\raggedright Hankel-DMD & \raggedright Target history divided by $B$. $W=\max\{16,\operatorname{round}(1.0f_s)\}$; delay $d=\max\{4,\operatorname{round}(0.040f_s)\}$. Rank at most 12; energy target 0.995; singular-value threshold $\max\{10^{-10}s_1,\epsilon\}$, with largest singular value $s_1$. & \raggedright Fit the truncated SVD at run onset, preferring fully reliable snapshot pairs. Propagate the delay-coordinate state during saturation.\tabularnewline[6pt]
\raggedright CR-RLS-ARX & \raggedright Same inputs, $p$, $d$, forgetting factor, initialization and normalization as RLS-ARX. Initial noise variance 0.04, floor 0.0009 and variance update rate $1-0.99^{200/f_s}$. & \raggedright Use observed updates at reliable samples and the censored-normal conditional mean at saturated samples. Update noise variance only at reliable samples.\tabularnewline[6pt]
\raggedright SSMKF-log-S & \raggedright Target-axis history with a normalized level/rate state and $\Delta t=1/f_s$. Initial state $(\widetilde z(0)/B,0)^{\mathsf T}$; $P=\operatorname{diag}(1,10)$; $\nu=3$. At most 30 iterations for each of two updates; tolerance $10^{-6}$. & \raggedright Filter observations at reliable samples. During saturation, predict the state and constrain endpoint sign and magnitude; write the constrained prediction back to the level state.\tabularnewline[6pt]
\end{longtable}\endgroup

SlopeGeo uses endpoint differences, scalar weights and a weighted median. For $M$ candidates, construction after a reliable history becomes available is $O(M)$, the current sort-based fusion is $O(M\log M)$ per evaluated sample, and storage is $O(H+M)$ for maximum history length $H$. Forward AR fits regression coefficients before each event. RLS-ARX and CR-RLS-ARX recursively update parameters and their covariance. Hankel-DMD constructs a delay embedding and performs a truncated singular-value decomposition. SSMKF-log-S propagates state and covariance and applies robust iterations. The history lengths and model orders in Tables~\ref{tab:orders} and \ref{tab:baselines} specify the computational workloads for runtime comparisons on a given platform.

The primary metric is pooled overrange RMSE normalized by each sequence's threshold,
\begin{equation}
\mathrm{RMSE}/B=\sqrt{\frac{1}{N}\sum_{j=1}^{N}\left(\frac{\hat z_j-z_j}{B_j}\right)^2},
\label{eq:pooled-normalized-rmse}
\end{equation}
where $z_j$ denotes the unclipped recorded or simulated reference, $B_j$ is its sequence-specific threshold, and the sum spans all evaluated overrange samples. The cumulative-error panels sum the signed normalized error from the start of the displayed window,
\begin{equation}
C(n)=\sum_{i=1}^{n}\frac{\widehat z_i-z_i}{B_i}.
\label{eq:cumulative-error}
\end{equation}
This dimensionless sum gives the net accumulated error within the window, allowing positive and negative errors to cancel. Computing angular displacement also requires time integration and conversion to physical units; computing three-dimensional attitude requires propagation of the rotational kinematics.

\section{Results and discussion}\label{sec:results}

We first compare pooled recovery error across all four data sources, then examine the waveform and record-level results for each source. The final analysis relates recovery error to saturation duration and local geometry using the controlled frequency--threshold sweep.

\subsection{Overall recovery performance}\label{sec:overall-results}

Table~\ref{tab:pooled} compares the seven methods using all evaluated overrange samples. Errors are normalized by each sequence's clipping threshold, so the reported values express recovery error relative to the imposed range boundary.

\begin{table}[!htbp]
\centering\small
\caption{Pooled overrange RMSE/$B$ for the seven methods. Bold indicates the lowest value in each column.}\label{tab:pooled}
\begin{tabular}{lrrrr}\toprule
Method & INSANE & UZH-FPV & Self-collected & Simulation\\\midrule
Clipped input & 1.5961 & 0.5289 & 0.6813 & 0.4799\\
SlopeGeo & \textbf{1.2279} & \textbf{0.4937} & \textbf{0.5408} & \textbf{0.2179}\\
Forward AR & 1.4536 & 0.5993 & 0.6757 & 0.2472\\
RLS-ARX & 1.5625 & 0.5209 & 0.5844 & 0.3797\\
Hankel-DMD & 1.4288 & 0.5922 & 0.7428 & 0.4027\\
CR-RLS-ARX & 1.3676 & 0.6164 & 0.5605 & 0.4180\\
SSMKF-log-S & 1.5293 & 0.5743 & 0.8683 & 0.3858\\
\bottomrule\end{tabular}
\end{table}
\FloatBarrier

SlopeGeo has the lowest pooled normalized error for each source, reducing error relative to clipped input by 23.1\% on INSANE, 6.7\% on UZH-FPV, 20.6\% on self-collected motion and 54.6\% in simulation. The closest recovery baseline differs across sources: CR-RLS-ARX on INSANE and self-collected motion, RLS-ARX on UZH-FPV and Forward AR in simulation. The following results examine these differences at the level of individual recordings and waveform windows.

\subsection{Recovery in INSANE flights}\label{sec:uav-results}

On INSANE, SlopeGeo reduced pooled overrange RMSE/$B$ from 1.596 for clipped input to 1.228, a 23.1\% reduction, and was 10.2\% lower than the closest recovery baseline, CR-RLS-ARX (1.368). Its mean RMSE across equally weighted recordings was 1.200 $\pm$ 0.489, compared with 1.586 $\pm$ 0.602 for clipped input and 1.321 $\pm$ 0.495 for CR-RLS-ARX (mean $\pm$ s.d.). 

SlopeGeo had lower file-level RMSE than clipped input in 26 of 26 recordings and than CR-RLS-ARX in 20 of 26. Both recovery methods use target-axis history and current auxiliary observations. Under these common input conditions, the results support the local geometric representation used by SlopeGeo in these recordings: a slope is estimated before each event, and displacement in the auxiliary plane tracks signal changes during clipping.

\begin{figure}[!htbp]
\centering
\includegraphics[width=0.94\textwidth,height=0.70\textheight,keepaspectratio]{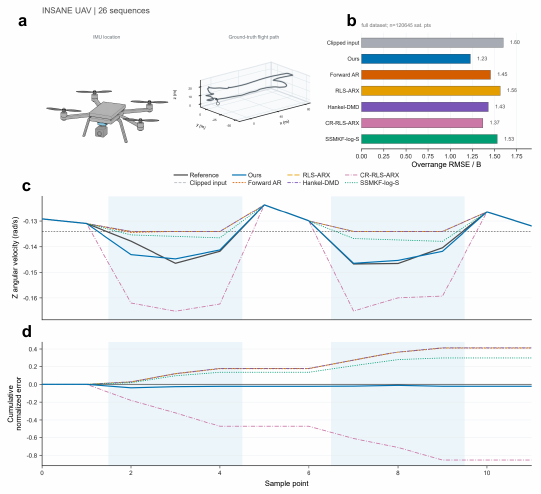}
\caption{\textbf{Strict-online recovery in INSANE flights.} \textbf{a,} Platform and trajectory context from outdoor\_1. \textbf{b,} Pooled overrange RMSE/$B$ across 26 sequences. \textbf{c,} A continuous recovery window from outdoor\_1. \textbf{d,} Cumulative signed error normalized by $B$ over the same window. SlopeGeo (Ours) uses the $Q=2$ final-output safeguard.}
\label{fig:insane_recovery}
\end{figure}

Figure~\ref{fig:insane_recovery}c shows two clipped negative troughs. SlopeGeo recovers their depth; the absolute cumulative normalized error at the end of the window is 0.022, compared with 0.410 for clipping (Fig.~\ref{fig:insane_recovery}d). The cumulative curve describes net angular-rate error in the target component, before propagation to three-dimensional attitude.

\subsection{Recovery in UZH-FPV racing flights}\label{sec:fpv-results}

\begin{figure}[!htbp]
\centering
\includegraphics[width=0.94\textwidth,height=0.70\textheight,keepaspectratio]{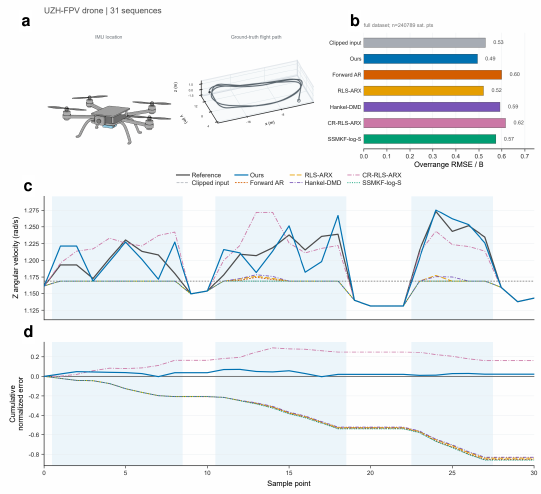}
\caption{\textbf{Strict-online recovery in UZH-FPV racing flights.} \textbf{a,} Platform and trajectory context from indoor\_45\_9\_snapdragon\_with\_gt. \textbf{b,} Pooled overrange RMSE/$B$ across 31 sequences. \textbf{c,} A continuous recovery window from the same recording. \textbf{d,} Cumulative signed error normalized by $B$. SlopeGeo (Ours) uses the same $Q=2$ final-output safeguard as in INSANE.}
\label{fig:fpv_recovery}
\end{figure}

Figure~\ref{fig:fpv_recovery} reports recovery during UZH-FPV racing flights. On this dataset, pooled RMSE/$B$ decreased from 0.529 to 0.494, a 6.7\% reduction, and was 5.2\% lower than the closest recovery baseline, RLS-ARX (0.521) (Fig.~\ref{fig:fpv_recovery}b). Mean RMSE across equally weighted recordings was 0.514 $\pm$ 0.131 for SlopeGeo, 0.559 $\pm$ 0.156 for clipped input and 0.538 $\pm$ 0.150 for RLS-ARX. SlopeGeo was lower than clipped input in 20 of 31 recordings and lower than RLS-ARX in 18 of 31. These recording-level statistics show how recovery gains vary across high-speed manoeuvres.

Rapid turns during racing flight can change the trajectory direction within a clipped peak, shortening the interval over which the pre-event slope remains applicable. Current auxiliary measurements constrain the missing waveform while this local relation holds. Differences in how long it holds may help explain the variation across UZH-FPV recordings. Sampling rate sets the temporal resolution at which these trajectory changes are observed. Figure~\ref{fig:fpv_recovery}c shows three positive peaks with different heights and shapes around their turning points. SlopeGeo preserves their relative heights and order, reducing the absolute cumulative normalized error at the end of the window from 0.855 to 0.022 (Fig.~\ref{fig:fpv_recovery}d). Recovering the rising and falling edges and peak amplitude reduces the signed error accumulated within each clipped peak.

\subsection{Recovery in self-collected inertial measurements}\label{sec:hardware-results}

\begin{figure}[!htbp]
\centering
\includegraphics[width=0.96\textwidth,height=0.60\textheight,keepaspectratio]{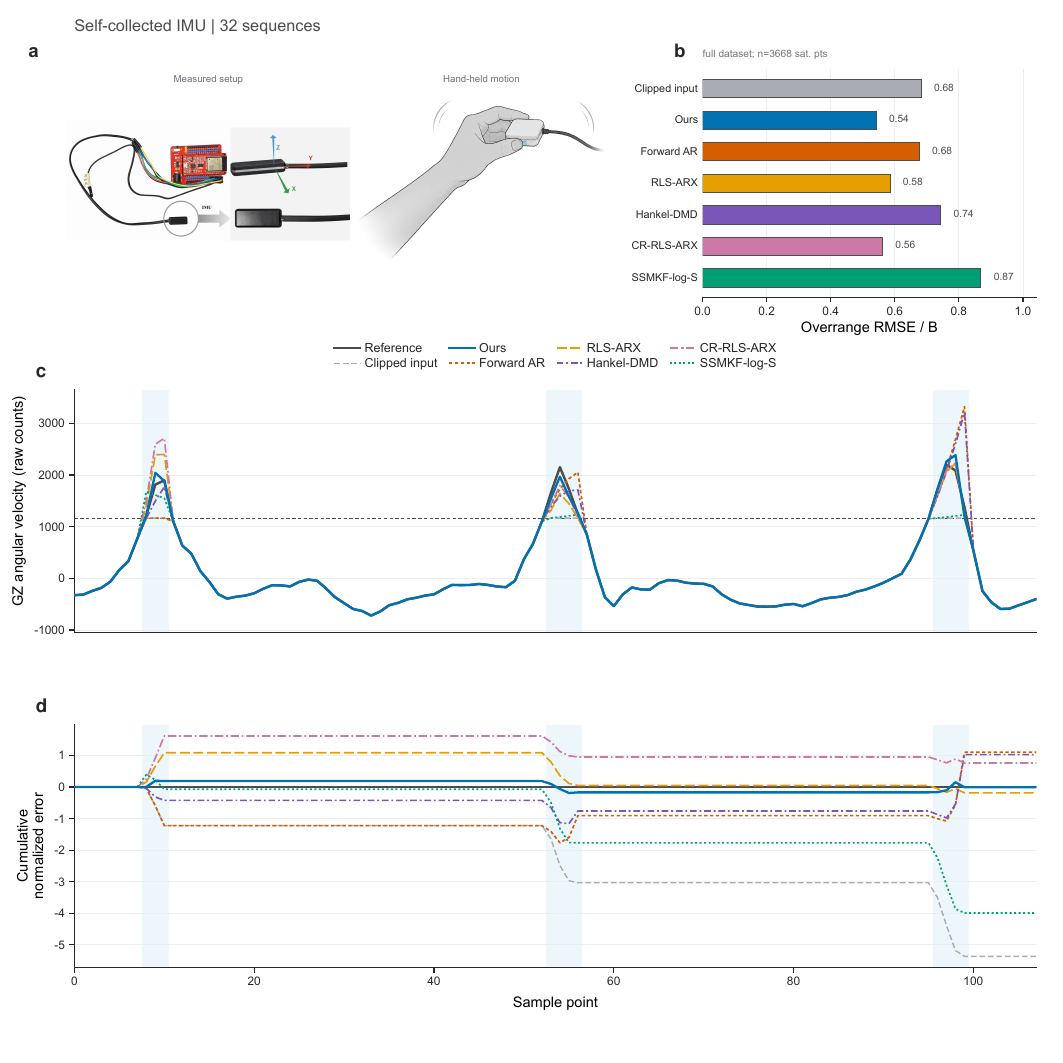}
\caption{\textbf{Recovery in self-collected human motion.} \textbf{a,} IMU setup and a schematic of hand-held motion. \textbf{b,} Pooled overrange RMSE/$B$ across 3,668 evaluated samples from 32 recordings. \textbf{c,} A 108-sample waveform example in the recorded numerical units, with three clipping runs shaded. \textbf{d,} Cumulative signed normalized error over the same window. All methods use the common $Q=2$ safeguard. Window selection and metric definitions are given in Sections~\ref{sec:validation} and \ref{sec:metrics}, respectively.}
\label{fig:selfcollected_recovery}
\end{figure}

Across the 32-record self-collected human-motion dataset, SlopeGeo reduced pooled overrange RMSE/$B$ from 0.681 for clipped input to 0.541, a reduction of 20.6\% across 3,668 evaluated samples (Fig.~\ref{fig:selfcollected_recovery}b). It achieved the lowest pooled error among the seven compared methods and was 3.5\% lower than the closest recovery baseline, CR-RLS-ARX (0.561). Mean RMSE across equally weighted recordings was 0.519 $\pm$ 0.271 for SlopeGeo, 0.648 $\pm$ 0.418 for clipped input and 0.524 $\pm$ 0.374 for CR-RLS-ARX.

Pooled RMSE averages the normalized squared errors over all evaluated samples and then takes the square root, whereas the recording-level mean averages the RMSE values calculated separately for each recording.

The type of motion may also affect recovery. During reciprocating or approximately planar motion, the relation between the dominant axis and the observable plane can remain valid around a brief peak, allowing Geo to recover its area. Faster direction changes in mixed-axis motion can shorten this interval. The 108-sample window in Fig.~\ref{fig:selfcollected_recovery}c contains three separated clipping runs. SlopeGeo gives a cumulative signed normalized error of $-0.008$ at the end of the window, with a maximum absolute excursion of 0.200; clipped input accumulates to $-5.367$ (Fig.~\ref{fig:selfcollected_recovery}d).

Each event uses the last reliable target measurement as its anchor and a slope estimated from the preceding reliable interval. Displacement in the auxiliary plane then guides recovery of the rising edge, peak and return to range. Once the target measurement is back within range, the output uses the measured value, so no further reconstruction error is added relative to the unclipped record.

\subsection{Recovery under controlled simulation}\label{sec:simulation-results}

\begin{figure}[!htbp]
\centering
\includegraphics[width=0.94\textwidth,height=0.70\textheight,keepaspectratio]{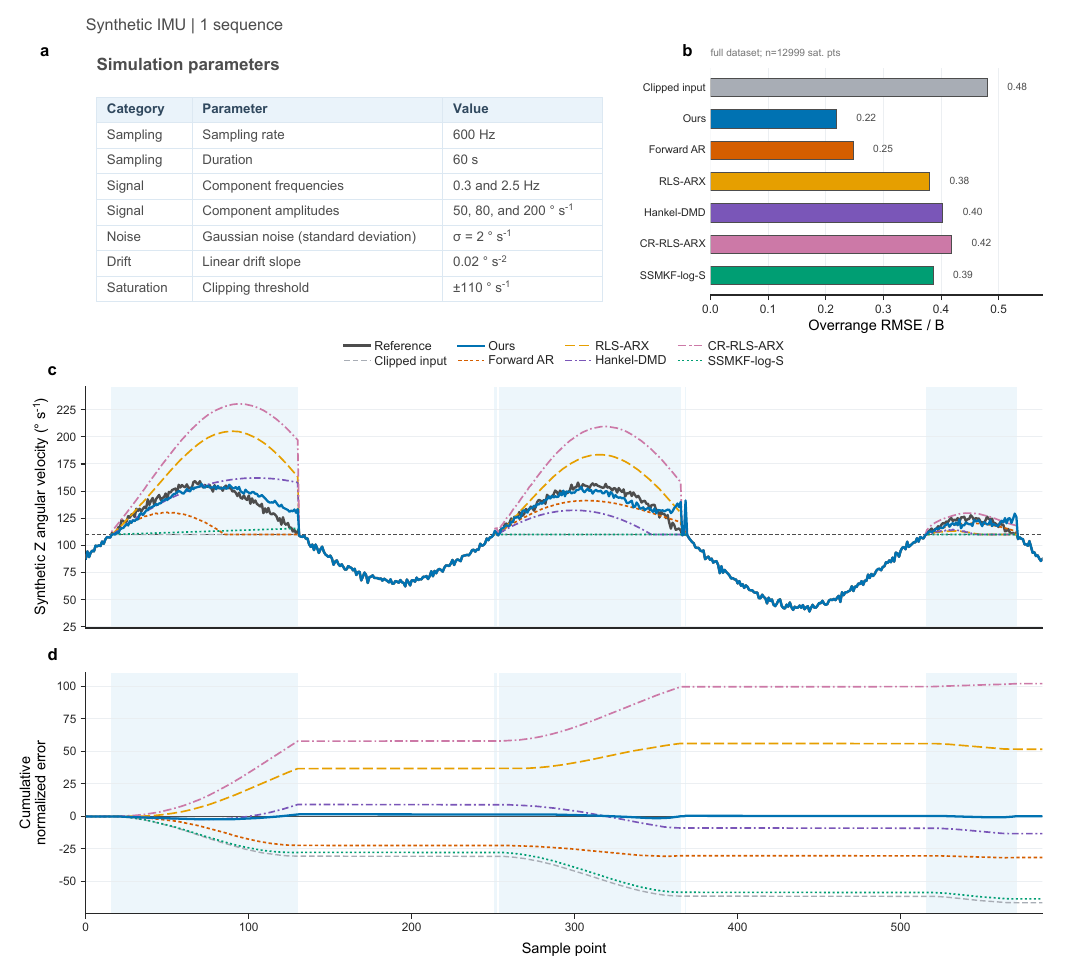}
\caption{\textbf{Recovery under controlled simulation.} \textbf{a,} Simulation settings; complete signal components, projection modulation and axis-specific drifts are specified in Section~\ref{sec:simulation-method}. \textbf{b,} Pooled overrange RMSE/$B$ over the 60-s sequence. \textbf{c,} A recovery waveform with three periodic clipping runs. \textbf{d,} Cumulative signed normalized error over the same window.}
\label{fig:simulation_boundary}
\end{figure}

The controlled simulation projects shared fast and slow components onto two angular-rate channels through a time-varying angle and uses a separate component for the third channel. Gaussian noise and axis-dependent drift are added to all three channels. Figure~\ref{fig:simulation_boundary}a summarizes the 60-s sequence defined in Section~\ref{sec:simulation-method}. Panels b--d show pooled overrange error, a recovery window and its cumulative error. The reference-assisted frequency--threshold sweep is examined separately in Section~\ref{sec:boundary-results}.

Across 12,999 evaluated overrange samples and 120 saturation events, SlopeGeo reduced pooled overrange RMSE/$B$ from 0.480 for clipped input to 0.218, a 54.6\% reduction (Fig.~\ref{fig:simulation_boundary}b), and was 11.9\% lower than the closest recovery baseline, Forward AR (0.247). This controlled case shows the largest improvement among the four sources, consistent with the cross-axis structure produced by shared components and a smoothly varying projection.

The periodic target waveform is favourable for Forward AR, the closest baseline in this experiment. As the projection changes, displacement in the observable plane reflects phase progression and allows Geo to recover the clipped peak. The window in Fig.~\ref{fig:simulation_boundary}c contains three periodic clipping runs. The reconstruction follows the broad peak envelope and preserves relative peak heights. After the three events, the absolute cumulative normalized error at the end of the window is 0.159 for SlopeGeo and 66.369 for clipped input (Fig.~\ref{fig:simulation_boundary}d).

A fixed cross-axis ratio loses accuracy as the projection changes. SlopeGeo selects among slopes from multiple history spans using pre-event validation weights. The smooth, broad peaks favour the use of those local slopes over a finite interval. The sweep below examines how event duration affects recovery.

\subsection{Effects of saturation duration and local geometry on recovery}\label{sec:boundary-results}

The output follows the measured target component between clipping events. Recovery within each event uses a slope estimated from reliable pre-event history. The local error analysis in Section~\ref{sec:boundary-method} separates the effects of slope variation, slope-estimation error and projected-path curvature, explaining why shorter events and stable local geometry favour accurate reconstruction.

\begin{figure}[!htbp]
\centering
\includegraphics[width=0.94\textwidth,height=0.70\textheight,keepaspectratio]{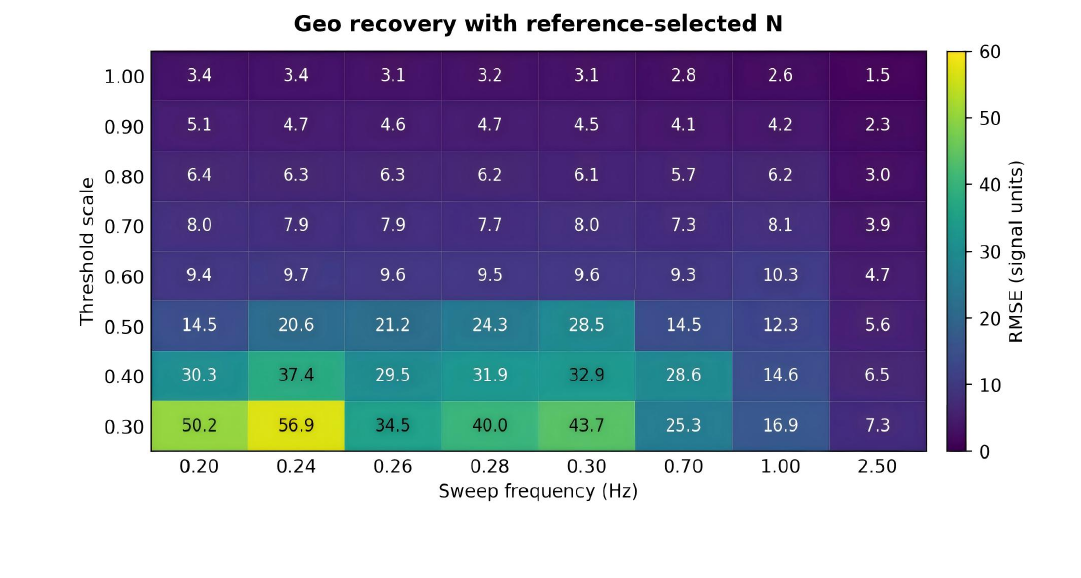}
\caption{\textbf{Reference-assisted Geo frequency--threshold sweep.} The reference signal is used to select the history span at each grid point. Full-sequence RMSE is averaged over four combinations of noise and drift. Colours and cell labels show error in the units of the sweep signal. This analysis evaluates geometric estimates selected using the reference signal; the online comparison is presented separately in Fig.~\ref{fig:simulation_boundary}.}
\label{fig:reference_sweep}
\end{figure}

In the reference-assisted sweep, we use the reference signal to select the best candidate history span at each grid point, then average the error over four combinations of noise and drift. At 0.24\,Hz, reducing the threshold scale from 0.50 to 0.30 increases mean full-sequence RMSE from 20.6 to 56.9 signal units; at scale 0.30, the value at 2.50\,Hz is 7.3 (Fig.~\ref{fig:reference_sweep}). The effect of frequency varies across the grid. Lowering the threshold increases both the amplitude to be recovered and the time spent above range. The error increase at 0.24\,Hz therefore reflects both a larger missing amplitude and a longer interval over which the pre-event relation must remain valid. At a fixed threshold scale, increasing frequency shortens each overrange interval. Projection reversal and the ratio of noise to displacement baseline also affect error. The interval over which recovery remains effective therefore depends on motion frequency, clipping depth, the curvature of the projected trajectory and sampling.

The error bound suggests three quantities for assessing recovery confidence: dispersion among candidate slopes, the projected baseline and changes in auxiliary trajectory direction. A short projected baseline makes the slope estimate more sensitive to noise, while trajectory reversal increases the difference between arc and chord lengths. The estimate returns to the measured target-axis value once the signal is back within range.

\FloatBarrier

Across the four data sources, SlopeGeo achieves the lowest pooled normalized error, while the closest baseline changes with the motion. The controlled simulation contains explicitly shared components and a smoothly varying projection, giving the largest recovery gain. The recorded motion and flight results show how the same Geo relation can be used under changes in direction, rhythm and sampling rate. The pre-event slope describes the local relation between components, while the current displacement in the auxiliary plane locates the signal trajectory relative to the recovery anchor. The continuing observations thus provide recovery information beyond the target-axis history. The error analysis explains why these gains depend on how long the local relation remains valid. Noise relative to the projected baseline affects slope estimation; path curvature and event duration affect the approximation during clipping. The observable projection and local slope can therefore guide future adaptive geometric recovery.

\section{Conclusions and outlook}\label{sec:discussion}\label{sec:conclusions}

This study presents Geo, a geometric model for recovering an angular-rate component during short-term saturation. Geo describes changes in the missing component along the trajectory in the observable plane, allowing measurements from the same IMU to support recovery of the saturated axis. Inertial continuity, a non-degenerate projection and local slope persistence define the conditions for this relation. The local error analysis connects recovery accuracy to slope variation, slope-estimation error and projected chord--arc discrepancy. The model and error analysis show how cross-axis information supports reconstruction and how local geometric changes affect its accuracy.

SlopeGeo translates Geo into an online algorithm using pre-event secant slopes and current auxiliary displacement. It achieved the lowest pooled normalized overrange RMSE among the seven methods in simulation, the 32 self-collected records and both flight datasets. Representative windows preserve peaks and valleys and reduce cumulative signed error.

Geo uses the spatial relation among measurements from existing sensing axes to recover waveforms within the same IMU. The effective recovery interval depends on how long the local relation remains valid. Small projected displacements make slope estimates more sensitive to noise, rapid trajectory changes increase approximation error, and saturation of the auxiliary axes reduces the available observations. The software-clipping experiments retain complete reference waveforms for paired evaluation. Assessing the device response during physical saturation will require synchronized measurements with a higher range.

Further work can use the geometric error terms to estimate recovery confidence and adapt the local model as motion changes. Hardware experiments with a higher-range reference could assess how sensor nonlinearity, hysteresis and return from saturation affect recovery. Using the recovered waveforms in attitude estimation and closed-loop control would allow us to quantify their effects on integrated motion error and system behaviour. Extending Geo to acceleration measurements or multiple saturated channels will require observable coordinates and local relations suited to those conditions.

\section*{Data availability}
HuGaDB, INSANE and UZH-FPV are available from their original dataset sources \cite{ref25,ref26,ref27}. Dataset details, comparison settings and aggregate results are provided in Section~\ref{sec:experimental-setup} and Table~\ref{tab:pooled}. The self-collected and simulated source recordings are available from the corresponding author on reasonable request.

\section*{Funding}
This work was supported by the Proof-of-Concept Project of Wuxi Institute of Industrial Technology (grant no. ZK202505026-JC-00[A]).

\section*{Competing interests}
The authors are pursuing patent protection relating to the Geo framework described in this study. The authors declare no other competing interests.

\section*{Declaration of generative AI and AI-assisted technologies in the manuscript preparation process}
The authors used ChatGPT for language editing, reviewed the resulting text and take full responsibility for the article.

\bibliographystyle{unsrtnat}
\bibliography{references}
\end{document}